\documentclass{article} 
\usepackage{iclr2026_conference,times}

\usepackage{amsmath,amsfonts,bm}

\def\eqref#1{equation~\ref{#1}}

\def\1{\bm{1}}

\DeclareMathAlphabet{\mathsfit}{\encodingdefault}{\sfdefault}{m}{sl}
\SetMathAlphabet{\mathsfit}{bold}{\encodingdefault}{\sfdefault}{bx}{n}

\usepackage{hyperref}
\usepackage{url}
\usepackage{graphicx}      
\usepackage{caption}       
\usepackage{subcaption}    
\usepackage{wrapfig}       
\usepackage{booktabs}
\usepackage{multirow}

\title{GeoEvolve: Automating Geospatial Model Discovery via Multi-Agent Large Language Models}

\iclrfinalcopy

\author{
Peng Luo \thanks{Equal contribution.} \\
Massachusetts Institute of Technology \\
\texttt{pengluo@mit.edu}
\And
Xiayin Lou \footnotemark[1]\\
Technical University of Munich  \\
\texttt{xiayin.lou@tum.de}
\And
Yu Zheng \\
Massachusetts Institute of Technology \\
\texttt{yu\_zheng@mit.edu}
\And
Zhuo Zheng, Stefano Ermon \\
Artificial Intelligence Laboratory \\
Stanford University \\
\texttt{\{zhzheng,ermon\}@stanford.edu}
}

\begin{document}

\maketitle

\begin{abstract}
Geospatial modeling provides critical solutions for pressing global challenges such as sustainability and climate change. Existing large language model (LLM)–based algorithm discovery frameworks, such as AlphaEvolve, excel at evolving generic code but lack the domain knowledge and multi-step reasoning required for complex geospatial problems. We introduce GeoEvolve, a multi-agent LLM framework that couples evolutionary search with geospatial domain knowledge to automatically design and refine geospatial algorithms. GeoEvolve operates in two nested loops: an inner loop leverages a code evolver to generate and mutate candidate solutions, while an outer agentic controller evaluates global elites and queries a GeoKnowRAG module—a structured geospatial knowledge base that injects theoretical priors from geography. This knowledge-guided evolution steers the search toward theoretically meaningful and computationally efficient algorithms. We evaluate GeoEvolve on two fundamental and classical tasks: spatial interpolation (kriging) and spatial uncertainty quantification (geospatial conformal prediction). Across these benchmarks, GeoEvolve automatically improves and discovers new algorithms, incorporating geospatial theory on top of classical models. It reduces spatial interpolation error (RMSE) by 13–21\% and enhances uncertainty estimation performance by 17 \%. Ablation studies confirm that domain-guided retrieval is essential for stable, high-quality evolution. These results demonstrate that GeoEvolve provides a scalable path toward automated, knowledge-driven geospatial modeling, opening new opportunities for trustworthy and efficient AI-for-Science discovery.
\end{abstract}

\section{Introduction}

Beyond building powerful AI models that help us analyze data and understand the world, enabling AI models to evolve on their own and autonomously extract knowledge stands as the next important and promising frontier. It usually involves a prolonged procedure of asking a research question, gathering relevant information, analyzing it to identify patterns or insights, and communicating the results as new knowledge. The rise of the large language models (LLMs), such as GPT-4 \citep{achiam2023gpt} and Gemini \citep{comanici2025gemini}, presents the possibility of accelerating and automating this knowledge discovery procedure. The confidence in this direction is supported by the breakthroughs in LLMs, such as retrieval augmented generation (RAG) that enhances the output of LLMs \citep{lewis2020retrieval,jiang2023active} and agents that execute complex tasks autonomously \citep{li2023camel,qian2024chatdev}. In fact, the integration of LLMs into this procedure has already boosted the performance of a range of discovery-oriented tasks, such as drug repurposing \citep{huang2024foundation}, hypothesis generation \citep{kumbhar2025hypothesis,xiong2024improving}, chip design \citep{ho2024large}, urban planning \citep{zhou2024large}. Recently, Google introduced AlphaEvolve, which has demonstrated remarkable capabilities in automating algorithm discovery across diverse domains, such as tackling complex mathematical optimization problems. Building on this foundation, OpenEvolve has been developed as an open-source implementation of Google DeepMind’s AlphaEvolve, providing the research community with accessible tools for further exploration and application.

Despite these advances, the domain of geospatial modeling remains relatively underexplored in the context of LLM-driven knowledge discovery. Geospatial problems are inherently complicated, characterized by spatial autocorrelation \citep{miller2004tobler}, spatial heterogeneity \citep{cheng2024ensemble}, scale effect \citep{chen2019quantifying}, and diverse modalities (e.g., maps, remote sensing imagery, spatial network, and textual description) \citep{mai2023opportunities}, etc. Moreover, addressing geospatial problems also demands synthesizing knowledge across different disciplines, from environmental science to urban studies, making it difficult for single-agent systems to provide comprehensive solutions.

\begin{figure}[t] 
  \centering
  \includegraphics[width=0.9\linewidth]{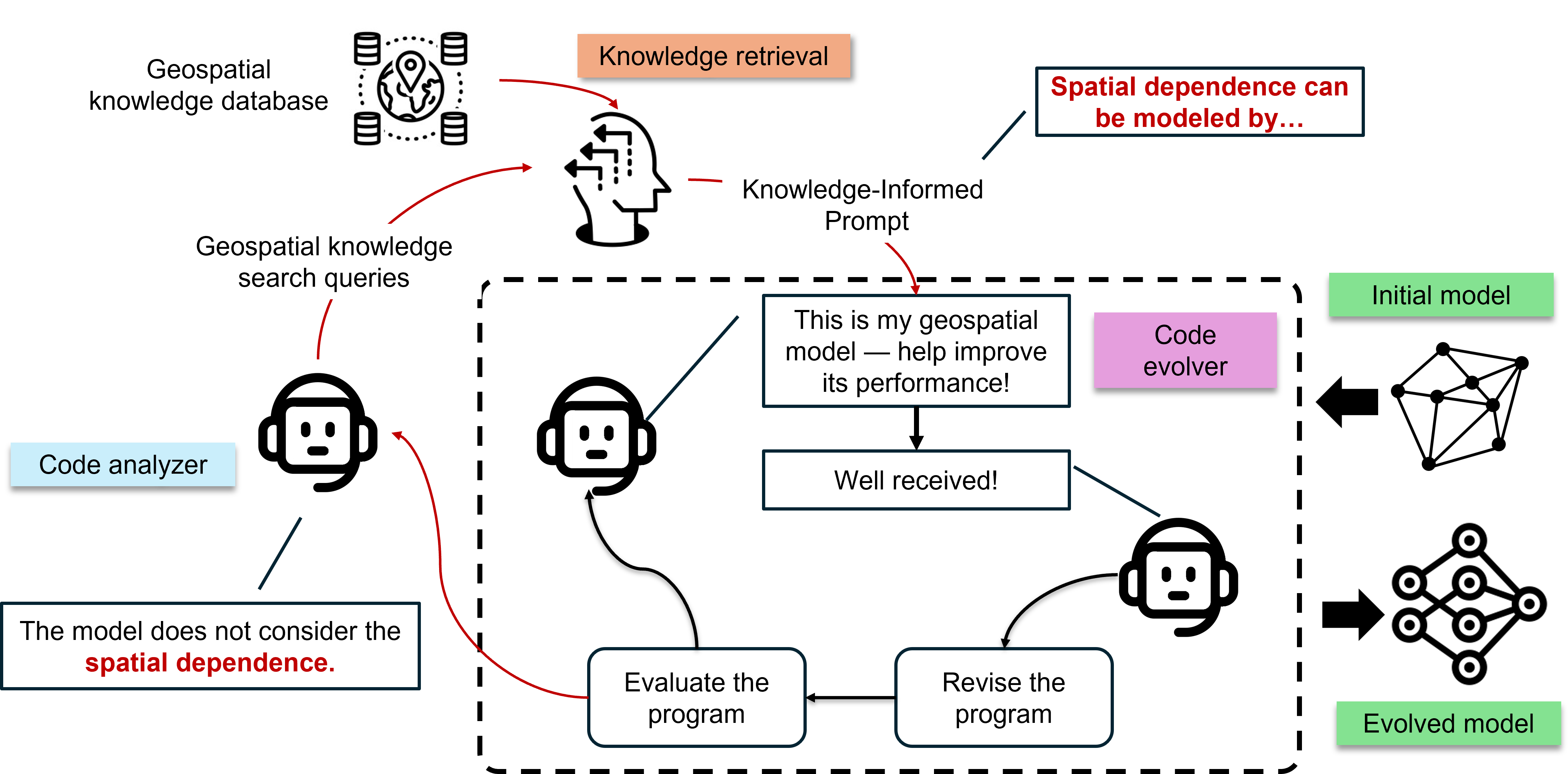}
  \caption{An illustration of the code-evolution trajectory of a geospatial model integrating domain knowledge. The dashed inner box represents the code evolver, a general algorithmic code-generation engine. The surrounding workflow depicts the knowledge-guided code generation proposed in this paper, specifically tailored for geospatial modeling.}
  \label{fig:illustration}
\end{figure}

In this paper, we introduce GeoEvolve, an advanced agent combining the evolutionary process with LLM-based code generation and geospatial knowledge-informed RAG (GeoKnowRAG) to automatically investigate optimal geospatial modeling. GeoEvolve operates in two complementary loops. As is shown in Figure \ref{fig:illustration}, the inner loop runs OpenEvolve \citep{openevolve} for a limited number of evolutionary steps, generating cadidates of discovery. The outer loop is governed by an agentic controller, which evaluates the best solutions, retains global elites to prevent performance degradation, and invokes the GeoKnowRAG module. This module will query a structured geospatial knowledge database, thus producing refined, domain-informed prompts that guide the next round evolution. We show that GeoEvolve can obviously improve the geospatial modeling. 

In summary, the contributions of our work are as follows:
\begin{enumerate}
  \item \textbf{Knowledge-guided evolution.}
  We integrate evolutionary search with domain knowledge by coupling GeoEvolve’s evolutionary code generation (via OpenEvolve) with retrieval-augmented geospatial knowledge. This grounds discovery in established geospatial theories and classical methods rather than random mutations, steering evolution toward theoretically meaningful and practically effective directions.
  \item \textbf{Automated, scalable pipeline.}
  We develop an automated and scalable geospatial modeling pipeline that can continuously evolve, adapt, and refine geospatial algorithms, providing a robust methodology for diverse geospatial tasks.
  \item \textbf{State-of-the-art performance and efficiency.}
  We demonstrate state-of-the-art performance on two spatial modeling cases—spatial interpolation and spatial uncertainty quantification—supported by an ablation study verifying the role of domain knowledge.
\end{enumerate}

\section{Related Work}
\paragraph{LLM-driven Algorithm Discovery}
Driven by LLMs, many studies aim to accelerate the discovery of algorithms with better performance, simpler implementation, and higher computational efficiency. A common approach is evolutionary search, which explores the algorithmic space via mutations and recombinations guided by performance metrics \citep{surina2025algorithm}, enabling breakthroughs across diverse applications \citep{lu2024discovering,ma2024eurekahumanlevelrewarddesign,velivckovic2024amplifying,morris2024llm}. Among the most influential methods is FunSearch—searching in the function space—which fosters creative algorithmic solutions while guarding against confabulations \citep{romera2024mathematical}, but is limited to evolving a single function rather than an entire codebase. AlphaEvolve, a substantially enhanced successor, leverages LLMs to solve complex problems at scale \citep{novikov2025alphaevolve}. Yet addressing specialized challenges, particularly in geospatial domains, requires domain-specific knowledge, multi-step reasoning, and iterative refinement guided by evaluation feedback \citep{chen2024llm}.

\paragraph{Retrieval-augmented generation RAG for scientific discovery.}
RAG has emerged as a standard strategy to ground LLM outputs in external knowledge, improving factual accuracy and controllability \citep{lewis2020retrieval,gao2023retrieval}.
Recent advances such as RAG-Fusion \citep{rackauckas2024rag} and reciprocal rank fusion (RRF) \citep{cormack2009reciprocal} demonstrate that expanding and fusing multiple reformulated queries can substantially enhance retrieval coverage and downstream reasoning quality.
Moreover, RAG has recently been applied in the geospatial domain to support knowledge discovery and contribute to downstream tasks such as spatial reasoning \citep{yu2025spatial}.
However, to the best of our knowledge, no prior work has leveraged RAG to extract geospatial knowledge specifically for geospatial model construction, leaving an important gap for integrating structured geographic knowledge into model design.


\section{GeoEvolve}
\begin{figure}[t] 
  \centering
  \includegraphics[width=0.8\linewidth]{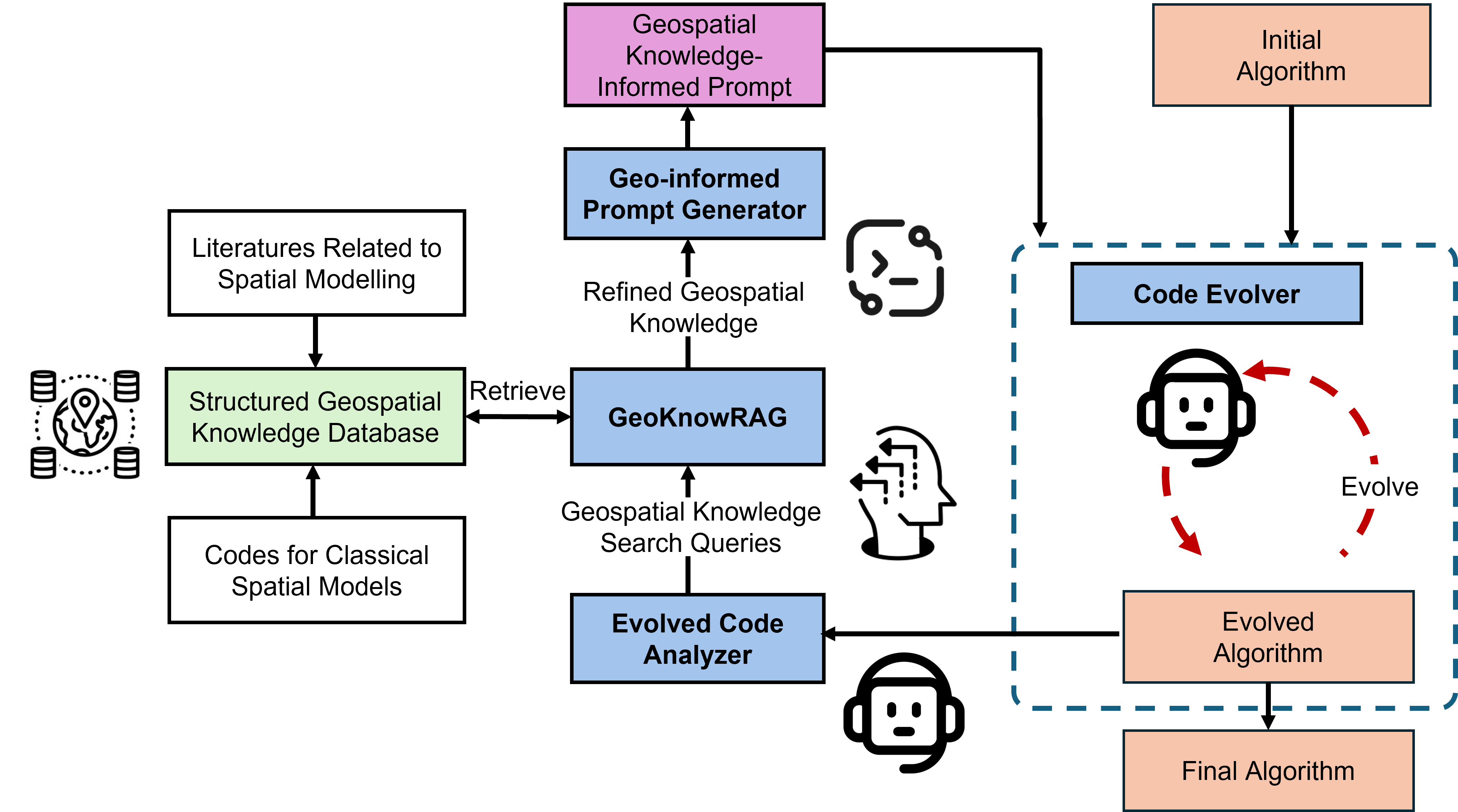}
  \caption{The workflow of GeoEvolve}
  \label{fig:geoevolve}
\end{figure}

GeoEvolve is designed to automate geospatial model discovery by integrating evolutionary code generation with structured geospatial knowledge. Unlike general-purpose code agents, GeoEvolve incorporates domain-specific knowledge from spatial modeling literature and classical algorithms, enabling the discovery of geospatial algorithms. Figure \ref{fig:geoevolve} illustrates the overall framework of GeoEvolve. It consists of four main components: (1) a code evolver, (2) an evolved code analyzer, (3) a geospatial knowledge retriever, and (4) a geo-informed prompt generator. Together, these components orchestrate a closed-loop process of code generation, evaluation, and refinement, leading to the emergence of geospatial model discovery.

\subsection{Code evolver}
The central engine of GeoEvolve is the code evolver, an evolutionary coding agent that generates and iteratively refines candidate algorithms. Beginning with an initial algorithm, the evolver performs a fully autonomous pipeline of mutation, evaluation, and selection relying on the power of LLMs. Candidate algorithms are represented as a group of executable code fragments. Mutations can be parameter changes, operator substitutions, or structure modifications to the algorithm. Abstractly, given a task-specific objective function $\mathcal{L}$, the evolver seeks to optimize an algorithm $A$ such that
\begin{equation}
    A*=\arg\min_{A\in \mathcal{A}}\mathcal{L}(A;\mathcal{D}),
\end{equation}
where $\mathcal{A}$ is the search space and $\mathcal{D}$ is the dataset. Here, we use OpenEvolve as the code evolver, which is the open-source equivalent of AlphaEvolve.

\subsection{Evolved code analyzer}
The evolved code analyzer is an LLM-powered diagnostic agent that interprets both the evolved code and associated metrics (e.g., RMSE for regression tasks). Its role is not limited to evaluating task outcomes, but also to providing semantic analysis of the code, thus identifying potential weaknesses or missing knowledge. To be specific, the LLM is required to achieve two tasks. First, it identifies missing or problematic knowledge from the evolved code. Second, it suggests search queries for retrieving useful geospatial knowledge from GeoKnowRAG. The diagnostic feedback given by this agent will be passed to the geospatial knowledge retriever to obtain related knowledge. This design allows GeoEvolve to reason about why the evolved algorithm fails and what kind of domain knowledge is needed to improve it. The template and an example of the code analyzer can be found at Figure \ref{fig:code_analyzer}.

\subsection{Geospatial knowledge retriever}

\begin{figure}[t] 
  \centering
  \includegraphics[width=0.8\linewidth]{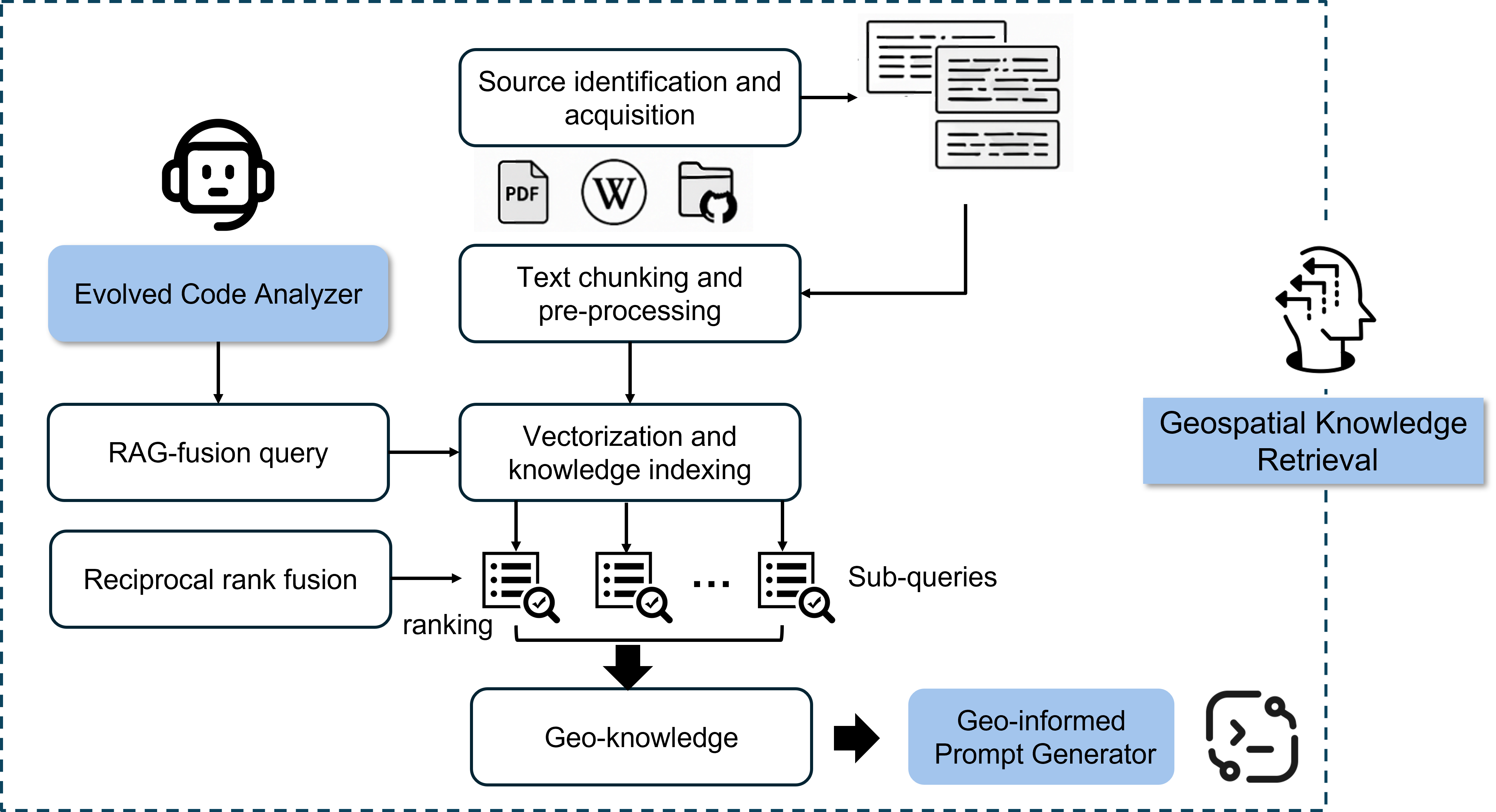}
  \caption{The workflow of GeoKnowRAG}
  \label{fig:rag}
\end{figure}

To prevent the evolutionary search from drifting into non-meaningful algorithmic space, GeoEvolve incorporates domain-specific geospatial knowledge through a dedicated Geospatial Knowledge Retrieval module (GeoKnowRAG). We construct a structured knowledge base by collecting literature on core geospatial modeling concepts (e.g., spatial autocorrelation) and classical algorithms (e.g., geographically weighted regression) from Wikipedia, arXiv, and GitHub, using curated keywords (Figure \ref{fig:database}, Appendix A.3.1). To ensure high-quality and comprehensive knowledge coverage, RAG-Fusion \citep{rackauckas2024rag} is applied to merge results from multiple reformulated queries, enabling the system to capture both precise theoretical matches and semantically related concepts. GeoKnowRAG transforms these diverse resources into a structured RAG system that delivers domain-aware prompts directly to the code evolver, providing the theoretical grounding and classical geospatial methods required for effective algorithmic refinement. As shown in Figure \ref{fig:rag}, GeoKnowRAG comprises four steps:

\paragraph{Source Identification and Acquisition}
First, we curate three complementary knowledge corpora: peer-reviewed geospatial modeling and algorithm papers in PDF form, authoritative encyclopedic entries from Wikipedia, and open-source code repositories from GitHub.  
Second, we compile a topic list that covers key spatial modeling concepts such as spatial autocorrelation, heterogeneity, kriging, geographically weighted regression, spatial conformal prediction, and network topology.  
Third, using these topics as queries, automated scripts call the Wikipedia, arXiv, and GitHub APIs to download relevant text and code, which are then converted into normalized UTF-8 \texttt{.txt} documents.

\paragraph{Text Chunking and Pre-processing}
First, each document is semantically segmented into 300-word chunks with a 50-word overlap to preserve contextual continuity across chunk boundaries and improve downstream retrieval accuracy.  
Second, all PDF, Markdown, and HTML sources are stripped of formatting, de-duplicated, and tokenized into a clean corpus ready for embedding.

\paragraph{Vectorization and Knowledge Indexing}
First, every chunk is encoded using the \texttt{text-embedding-3-small} model from OpenAI to obtain high-dimensional semantic vectors.  
Second, these embeddings are stored in a \textbf{Chroma} vector database, which supports approximate nearest-neighbor search and metadata filtering by topic or source type.  
Third, this indexed database forms the persistent memory of GeoKnowRAG and enables millisecond-scale retrieval across the geospatial knowledge space.

\paragraph{RAG-Fusion Query and Prompt Generation}
First, GeoKnowRAG employs multi-angle question expansion, where each input query from the GeoEvolve controller is reformulated into several sub-questions emphasizing different semantic aspects such as theory, implementation, and evaluation.  
Second, each sub-question is independently embedded and used for vector search to retrieve top-$k$ relevant chunks from the Chroma index.  
Third, the retrieved results are re-ranked using RRF, which scores passages based on the reciprocal of their ranks across sub-queries so that consistently high-scoring chunks surface to the top.  
Fourth, the highest-ranked passages are aggregated and summarized into a geo-informed prompt encoding key formulas, algorithmic structures, and empirical heuristics, which is then supplied to the GeoEvolve code evolver to guide the next round of algorithmic mutation and evaluation.

\subsection{Geo-informed prompt generator}
The information, from retrieved geospatial knowledge to evolved code, and associated metrics, is then processed together by the geo-informed prompt generator, which will translate it into a structured prompt for the code evolver. This prompt refines the search by introducing domain constraints, suggesting algorithmic structures, or incorporating empirical heuristics. The generator leverages LLMs as reasoning and translation engines, transforming abstract geospatial knowledge into actionable modifications of candidate code. 

The LLMs are required to generate a prompt that includes four key elements. 
First, algorithmic fixes or improvements suggesting how the current algorithm could be revised. 
Second, new operators or parameters that may improve performance in subsequent evolutionary iterations. 
Third, geospatial knowledge, including the direction of exploration, theoretical or empirical conditions, and expected outputs. 
Fourth, maximum tokens control, which helps maintain efficiency and reduce hallucination.

\section{Experiments}

To evaluate GeoEvolve’s capability for improving and discovering geospatial models, we focus on two fundamental topics in geospatial modeling: spatial interpolation and uncertainty quantification of spatial prediction.
For each topic, we select the most representative and classical baseline model, and employ a GPT-4–based evolutionary engine as the core evolve agent to autonomously search, mutate, and refine candidate algorithms.

We use OpenEvolve as the primary baseline. In addition, we conduct an ablation study with two variants. 
First, OpenEvolve with GeoKnowledge Prompt, where domain knowledge is incorporated as additional prompts. 
The prompt template is: 
\emph{``You are allowed to refer to advanced methods in the field of spatial interpolation and consider some important settings of spatial models, such as localized variogram, automatic variogram parameter selection, or stratified strategy, etc.''} Second, GeoEvolve without GeoKnowledge, where the GeoKnowRAG module is removed. For every algorithm, after each evolutionary step the generated code is first analyzed by the code analyzer and then directly passed to the knowledge-prompt generator to create new prompts.

For the OpenEvolve-based algorithms, we perform ten iterations of evolutionary search. For the GeoEvolve algorithms, we run ten outer-loop cycles—each consisting of the code analyzer, GeoKnowRAG, and geo-informed prompt generator—and within every outer cycle we conduct ten inner-loop evolutions. This results in a total of one hundred evolutionary iterations. For every experiment, the dataset is split into training, validation, and test sets in an 8:1:1 ratio.

\subsection{Spatial interpolation model}

\paragraph{Task- Spatial interpolation}
Spatial interpolation is one of the most important applications in geospatial analysis and a key approach for humans to observe the Earth’s surface environment and understand the planet \citep{lam1983spatial}.
Its task is to model discrete sample points collected across geographic space—such as climate observation stations, biodiversity observation points, or mineral sampling sites—and to predict the continuous spatial surface of the geographic variables of interest based on these observations.

\paragraph{Model- Oridinary Kriging}
We selected ordinary kriging, the most classical geostatistical spatial interpolation model, as the first case study for GeoEvolve to automatically improve and evaluate.
Since its invention, many studies have attempted to extend kriging, for example by integrating regression models in regression kriging \citep{hengl2007regression} or by accounting for spatially stratified heterogeneity in stratified kriging \citep{luo2023generalized}. However, ordinary kriging remains the fundamental core of the entire kriging family and of geostatistics itself.
Because it was developed long ago and has a relatively simple structure, direct algorithmic innovations to ordinary kriging have become increasingly rare. More details about ordinary kriging can be found at Appendix A.3.1.

If GeoEvolve can demonstrably enhance ordinary kriging, it would greatly revitalize geostatistical methods and provide fundamental improvements that can propagate to all kriging-based models and applications.This rationale underpins our choice of ordinary kriging as the first benchmark algorithm in this study. 

\paragraph{Evaluator}For the kriging interpolation task, we use the root mean squared error (RMSE) as the evaluation metric. 
Our objective is to obtain a kriging model that achieves a lower RMSE, indicating higher predictive accuracy.

\paragraph{Datasets}
In this study, we use trace-element observations of copper (Cu), lead (Pb), and zinc (Zn) collected from a representative region of Australia (with concentrations expressed in parts per million, ppm) to conduct spatial interpolation and geostatistical modeling experiments.
These three heavy metals have important indicative significance in environmental geochemistry: on the one hand, they serve as key factors for assessing regional environmental pollution levels and soil heavy-metal accumulation.
Details of the data acquisition and processing procedures can be found in \citep{luo2025feature}.

\subsubsection{Evolved algorithm of ordinary kriging.}

GeoEvolve preserves the ordinary-kriging core but augments it with (i) an expanded variogram family (Exponential, Gaussian, Linear, and Matérn) with automatic model selection via AIC/BIC, capturing a wider range of spatial smoothness; (ii) an adaptive empirical variogram using quantile/Silverman binning, trimmed means, and an automatic $n_{\text{lags}}\!\in\![8,20]\propto\sqrt{n}$ to stabilize nugget/sill/range estimation; (iii) robust multi-start fitting with L1 or weighted least squares and bin-based weights to avoid local minima and keep parameters physically meaningful; (iv) localized kriging that solves a $K$-NN system with condition-number–aware diagonal adjustment, reducing complexity from $O(n^3)$ to $O(K^3)$ and improving numerical stability; and (v) an adaptive log transform with a data-driven offset to reduce skew and ensure valid back-transformation. Together, these changes retain unbiasedness and best-linear prediction while delivering lower RMSE/MAE, tighter residuals, and greater computational robustness across heterogeneous spatial settings. The detailed development of GeoEvolve--Kriging can be found at Appendix A.4.1.

\subsubsection{Model evaluation}

Table \ref{tab:geo-evolve-metrics} reports the kriging accuracy obtained by different methods.  
GeoEvolve--kriging consistently achieves the lowest RMSE and MAE across the prediction of Cu, Pb, and Zn, while the original kriging baseline performs worst.  
Applying OpenEvolve to kriging improves the prediction of Cu and Pb but slightly degrades the performance on Zn.  
Introducing GeoKnowledge prompts into OpenEvolve does not lead to further gains, possibly because the injected knowledge lacks direct relevance to variogram estimation or spatial covariance structures that govern kriging performance.  
GeoEvolve without GeoKnowRAG already outperforms OpenEvolve, yet still falls short of the full GeoEvolve model, underscoring the critical role of structured geospatial domain knowledge in guiding algorithm evolution.

Compared with OpenEvolve--kriging, GeoEvolve--kriging reduces RMSE by 11.3\%, 20.9\%, and 13.5\% on Cu, Pb, and Zn predictions, respectively.  
Relative to the original kriging, the reductions are 15.4\%, 21.2\%, and 13.0\%, further highlighting GeoEvolve's ability to automatically discover and refine spatial interpolation algorithms with substantially improved predictive accuracy.

\begin{table*}[t]
\scriptsize
\centering
\caption{Performance comparison across different methods. For each metal, lower is better for RMSE/MAE, and higher is better for $R^2$.}
\label{tab:geo-evolve-metrics}
\begin{tabular}{@{}l ccc ccc ccc@{}}
\toprule
\multirow{2}{*}{\textbf{Method}} & \multicolumn{3}{c}{\textbf{Cu}} & \multicolumn{3}{c}{\textbf{Pb}} & \multicolumn{3}{c}{\textbf{Zn}} \\
\cmidrule(lr){2-4} \cmidrule(lr){5-7} \cmidrule(lr){8-10}
 & \textbf{RMSE} $\downarrow$ & \textbf{MAE} $\downarrow$ & $\mathbf{R^2}$ $\uparrow$ & \textbf{RMSE} $\downarrow$ & \textbf{MAE} $\downarrow$ & $\mathbf{R^2}$ $\uparrow$ & \textbf{RMSE} $\downarrow$ & \textbf{MAE} $\downarrow$ & $\mathbf{R^2}$ $\uparrow$ \\
\midrule
Original & 0.9348 & 0.6841 & 0.3128 & 0.6752 & 0.4666 & 0.2657 & 0.6520 & 0.4806 & 0.3634 \\
\addlinespace
OpenEvolve (Baseline) & 0.8919 & 0.6740 & 0.3776 & 0.6723 & 0.5063 & 0.2781 & 0.6561 & 0.5050 & 0.3583 \\
\addlinespace
OpenEvolve with GeoKnowledge & 0.9349 & 0.6842 & 0.3126 & 0.6760 & 0.4674 & 0.2639 & 0.6520 & 0.4808 & 0.3635 \\
\addlinespace
GeoEvolve without GeoKnowRAG & 0.8308 & 0.6146 & 0.3706 & 0.5709 & 0.4075 & 0.3240 & 0.6092 & 0.4549 & \bf{0.3978} \\
\addlinespace
GeoEvolve & \bf{0.7910} & \bf{0.6001} & \bf{0.3896} & \bf{0.5320} & \bf{0.3935} & \bf{0.3792} & \bf{0.5672} & \bf{0.4493} & 0.3080 \\
\bottomrule
\end{tabular}
\end{table*}

\begin{figure}[t] 
  \centering
  \includegraphics[width=0.8\linewidth]{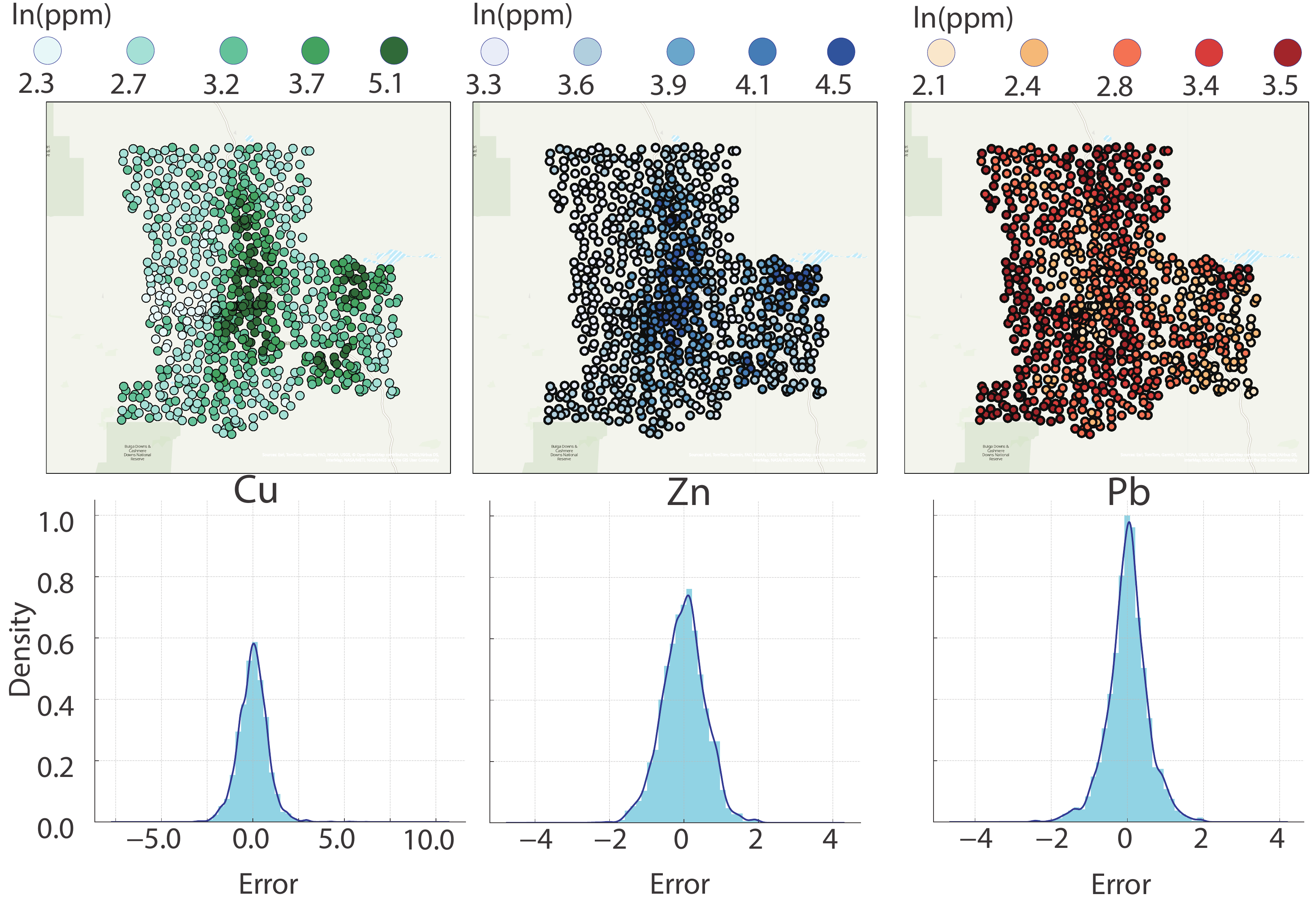}
  \caption{The spatial distribution of predicted concentrations and the error distribution of three elements, Cu, Zn, and Pb obtained from Evolved Kriging}
  \label{fig:map_kriging}
\end{figure}

Figure~\ref{fig:map_kriging} illustrates the spatial distributions of the predicted concentrations and the associated error maps for Cu, Pb, and Zn obtained by GeoEvolve--kriging, clearly demonstrating its capability to capture fine-scale spatial variability while maintaining low residual errors.

\subsection{Spatial uncertainty quantification model}

\paragraph{Task- Spatial UQ}

In spatial predictive modeling, it is not sufficient merely to develop more accurate models for point predictions; an equally critical task is to quantify and communicate the uncertainty of predictions, as this directly shapes the reliability and legitimacy of geography-based decisions such as flood evacuation planning and public facility site selection. Therefore, incorporating rigorous uncertainty quantification into spatial prediction is essential not only for improving scientific credibility, but also for supporting transparent, fair, and ethically sound spatial planning and policy making.



\paragraph{Model- GeoCP}
In geography, the task of assessing the reliability of spatial prediction results is commonly addressed through uncertainty quantification (UQ).
In this study, we adopt geospatial conformal prediction (GeoCP)—a model-agnostic algorithm for estimating the uncertainty of spatial prediction models—as the target method for enhancement using GeoEvolve \citep{lou2025geoconformal}. More details about GeoCP can be found at Appendix A.3.2.

\paragraph{Evaluator}
For GeoCP uncertainty estimation, we use the \emph{interval score}  
\begin{equation}
\text{IS}_i=\max(U_i-L_i,\epsilon)+\frac{2}{\alpha}\big[(L_i-y_i)\mathbb{I}(y_i<L_i)+(y_i-U_i)\mathbb{I}(y_i>U_i)\big],
\end{equation}
where $L_i,U_i$ are prediction bounds, $y_i$ the observation, and $\alpha$ the significance level (e.g., $0.1$ for 90\% intervals).  
The first term measures interval width (with $\epsilon\!\approx\!10^{-6}$ to avoid zero width), and the second penalizes coverage violations, scaled by $1/\alpha$.  
Smaller $\text{IS}$ indicates tighter and better-calibrated intervals.

\paragraph{Datasets}

The housing price dataset used in this study originates from the GeoDa Lab repository\footnote{\href{https://geodacenter.github.io/data-and-lab/KingCounty-HouseSales2015/}{https://geodacenter.github.io/data-and-lab/KingCounty-HouseSales2015/}}. 
The original data include 21,613 residential transactions and 21 attributes from Seattle and King County, Washington (May 2014–May 2015). 
For our analysis, we focus on the Greater Seattle urban core and retain 11 key variables, with housing sale price (in \$10,000s) as the dependent variable. 
Eight non-spatial predictors capture structural and quality characteristics—bathrooms, living-space and lot size, grade, condition, waterfront proximity, view quality, and property age—while two spatial predictors are geographic coordinates expressed in UTM. 
Further details of the dataset are documented in \citep{lou2025geoconformal}.

\subsubsection{Evolved algorithm of GeoCP}


GeoEvolve–GeoCP preserves the fundamental conformal prediction framework of GeoCP while introducing two major methodological advances. First, it refines the geographic weighting scheme: still employing a Gaussian kernel, but re-optimizing the bandwidth parameter through multi-start global search with adaptive clipping to ensure numerical stability and faithfully capture local spatial heterogeneity. Second, it enhances the weighted quantile computation by unifying earlier adaptive strategies into a simplified yet robust stepwise estimator with improved vectorization and conditioning checks, thereby delivering higher accuracy and better scalability on large test sets.The detailed analysis of GeoEvolve-GeoCP can be found at Appendix A.4.2.

\subsubsection{Model evaluation}

\begin{figure}[t] 
  \centering
  \includegraphics[width=0.8\linewidth]{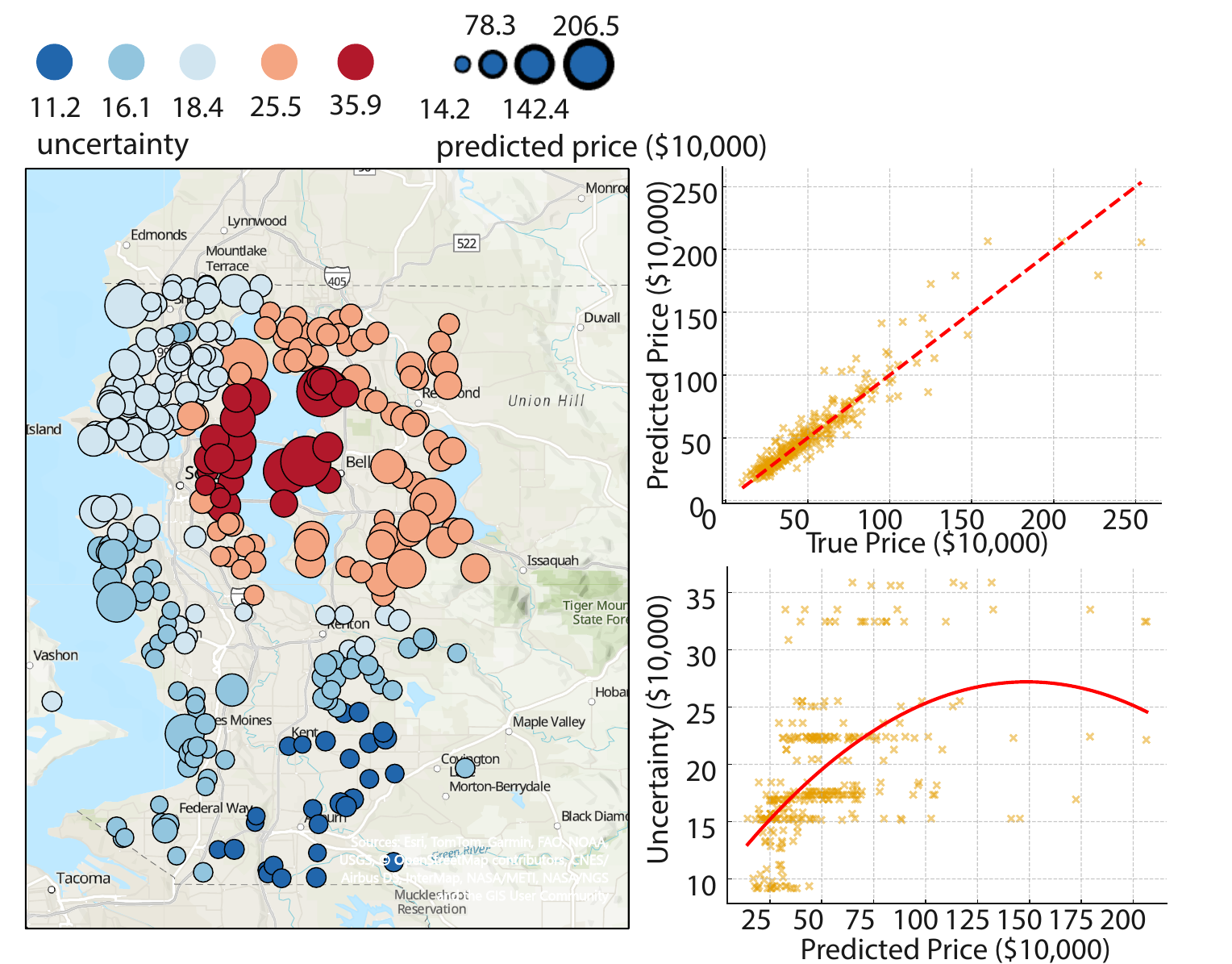}
  \caption{The spatial distribution of estimated uncertainty for houseing price prediction task in Seattle using the evolved GeoCP.}
  \label{fig:map_geocp}
\end{figure}

To perform GeoCP, we first build a house-price prediction model using a base predictor with eight explanatory variables and two spatial variables as inputs. The trained model is then assessed with GeoCP to quantify predictive uncertainty, and the final output is the uncertainty of house-price predictions on the test set. In this study, we choose XGBoost as the base predictor, which achieves an $R^2$ of 0.871 and an RMSE of 7.362 (10,000~USD).  The results are presented in Figure \ref{fig:map_geocp}. The predicted uncertainty exhibits a clear spatial pattern: it is highest around Lake Washington in downtown Seattle, slightly lower in suburban areas, and lowest in the rural southern region. A scatter plot of predicted uncertainty versus predicted price further reveals that uncertainty increases with house price, peaking at approximately 125 (10,000~USD) and then leveling off with a slight decline.

We apply GeoCP in four configurations, original, OpenEvolve, OpenEvolve with GeoKnowledge Prompt, and GeoEvolve without GeoKnowRAG--to quantify uncertainty on the same test set. 
Table \ref{tab:interval-metrics} reports the GeoCP performance obtained by different methods. As shown, OpenEvolve reduces the interval score from 55.37 to 52.37. 
Adding GeoKnowledge to OpenEvolve does not further improve performance. 
In contrast, GeoEvolve achieves an interval score of 46.12, representing reductions of 16.7\% and 11.9\% compared with the original GeoCP and OpenEvolve–GeoCP, respectively.

\begin{table*}[t]
\centering
\caption{Comparison of conformal prediction metrics. 
Smaller Average Interval Size and Interval Score indicate sharper and more efficient intervals.}
\label{tab:interval-metrics}
\begin{tabular}{@{}lcc@{}}
\toprule
\textbf{Method} & \textbf{Average Interval Size} $\downarrow$ & \textbf{Interval Score} $\downarrow$ \\
\midrule
Original & 19.9471 & 55.3692 \\
\addlinespace
OpenEvolve (Baseline) & 16.0036 & 52.3705 \\
\addlinespace
OpenEvolve with GeoKnowledge & 19.2948 & 54.7979 \\
\addlinespace
GeoEvolve without GeoKnowRAG & 16.8182 & 50.5407 \\
\addlinespace
GeoEvolve & \bf{12.0461} & \bf{46.1195} \\
\bottomrule
\end{tabular}
\end{table*}

\section{Conclusion}
We presented GeoEvolve, a multi-agent LLM framework that couples evolutionary code search with geospatial domain knowledge via GeoKnowRAG to automate geospatial model discovery. Across two fundamental tasks—spatial interpolation (ordinary kriging) and spatial uncertainty quantification (GeoCP)—GeoEvolve consistently improved upon classical baselines and strong OpenEvolve variants. Ablations confirm that structured, domain-guided retrieval is pivotal: removing GeoKnowRAG degrades performance despite identical evolutionary budgets, underscoring the value of grounding algorithm evolution in geospatial theory. In the future, we plan to evaluate the performance of GeoEvolve with different foundation models and to incorporate a broader and more comprehensive geospatial knowledge database.

\clearpage
\bibliography{iclr2026_conference}
\bibliographystyle{iclr2026_conference}

\appendix
\section{Appendix}

\subsection{Use of LLMs}
We use LLMs to polish selected paragraphs and to automatically extract differences between algorithms (e.g., Kriging and GeoCP) produced by different code-generation methods (e.g., OpenEvolve and GeoEvolve), thereby facilitating the analysis of GeoEvolve’s specific improvements and their underlying causes. All research ideas were independently conceived by the authors.

\subsection{Code analyzer}

\begin{figure}[t] 
  \centering
  \includegraphics[width=0.9\linewidth]{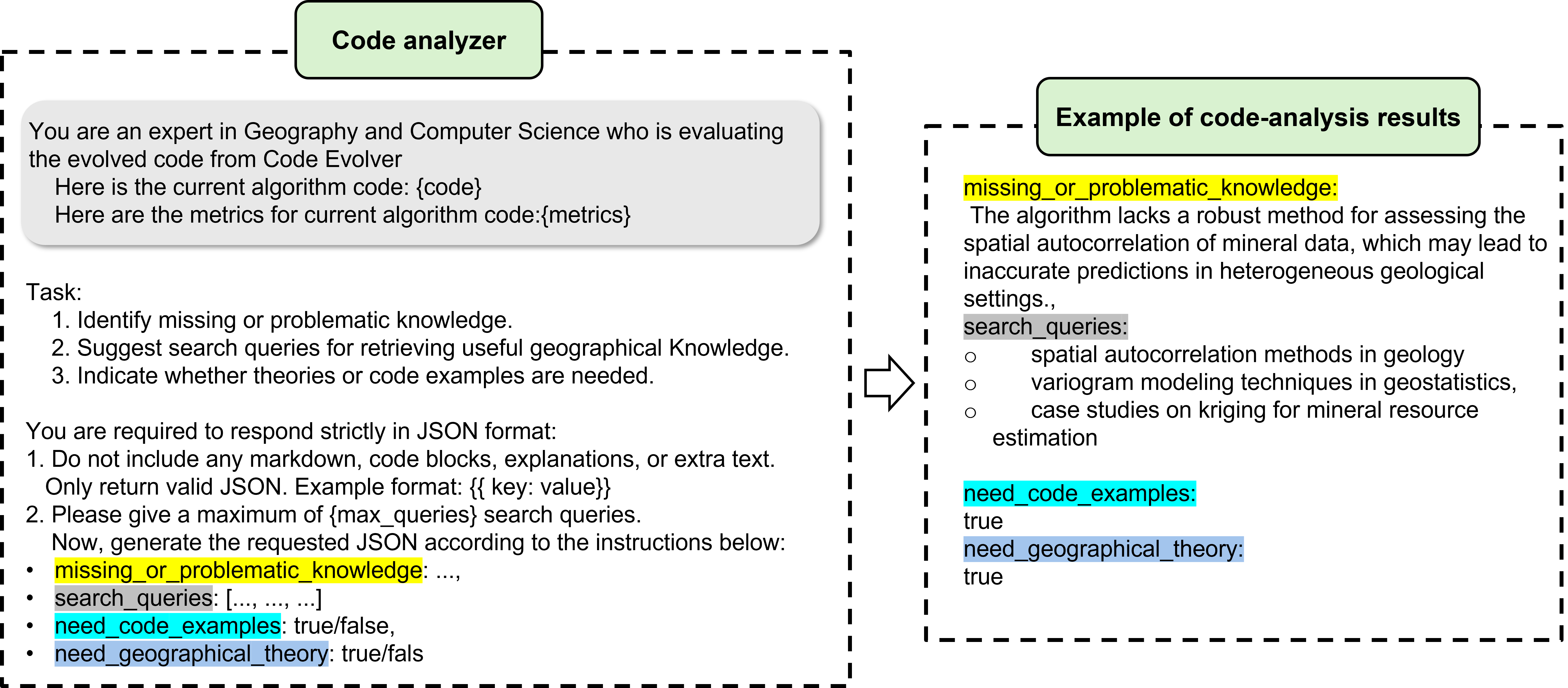}
  \caption{The template and an example of code analyzer}
  \label{fig:code_analyzer}
\end{figure}

Figure \ref{fig:code_analyzer} shows the template of the Code Analyzer and an example output.

\subsection{Geospatial knowledge database}

\begin{figure}[t] 
  \centering
  \includegraphics[width=0.9\linewidth]{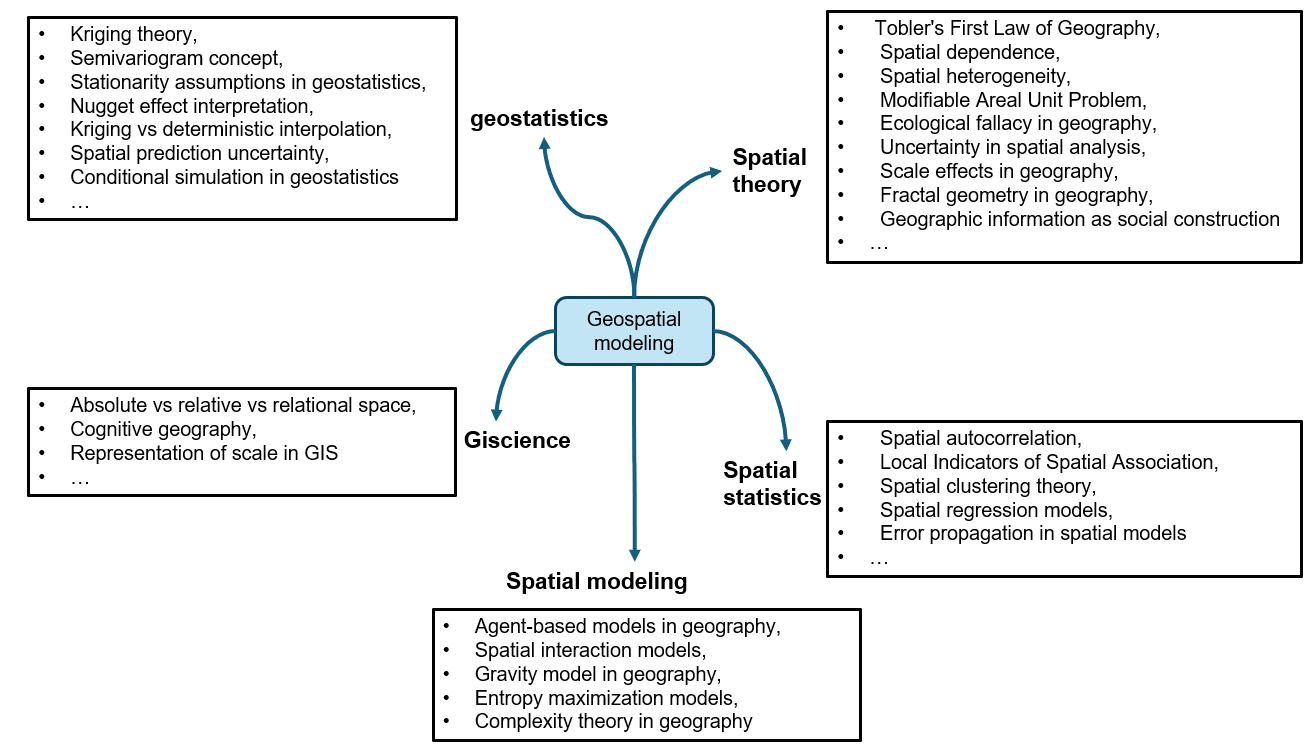}
  \caption{The keywords used for constructing geospatial knowledge database}
  \label{fig:database}
\end{figure}

In this study, we retrieved geospatial knowledge using five categories of keywords, including publications, GitHub code, and Wikipedia. As is shown in Figure \ref{fig:database}, the five categories are geostatistics, spatial theory, GIScience, spatial statistics, and spatial modeling.
The keywords within each category were proposed by the authors based on their domain expertise, reflecting concepts we consider particularly important for geospatial modeling.
In total, 141 knowledge documents were constructed.

It should be noted that the construction of a geospatial knowledge base can include many more keywords, enabling a much larger scale—potentially comprising thousands of documents or developed through more sophisticated processes.
In the present experiments, however, we intentionally created a small-scale knowledge base to validate the effectiveness of GeoEvolve on two algorithmic tasks.
We expect that GeoEvolve will achieve even greater performance gains when combined with a larger and more comprehensive geospatial knowledge base in future work.

\subsection{Original algorithm}

\subsubsection{Original algorithm of oridinary kriging.}
Kriging is a geostatistical spatial interpolation method that provides the \emph{best linear unbiased estimator} (BLUE) of an unknown value at a location by optimally weighting surrounding observations.
It assumes that the spatial process $Z(s)$ can be represented as
\begin{equation}
    Z(s) = \mu + \varepsilon(s),
\end{equation}
where $\mu$ is an unknown constant mean and $\varepsilon(s)$ is a zero-mean, second-order stationary random field.  
The key assumption of second-order stationarity requires that the mean is constant and that the covariance depends only on the lag vector $h$, i.e.,
\begin{equation}
    \operatorname{Cov}\big[ Z(s), Z(s+h) \big] = C(h),
\end{equation}
or equivalently through the semivariogram $\gamma(h)$.

Ordinary kriging predicts the value at an unsampled location $s_0$ as a weighted linear combination of the observed data:
\begin{equation}
    \hat{Z}(s_0) = \sum_{i=1}^n \lambda_i Z(s_i),
\end{equation}
subject to the unbiasedness constraint
\begin{equation}
    \sum_{i=1}^n \lambda_i = 1.
\end{equation}
The kriging weights $\lambda_i$ are determined by minimizing the estimation variance
\begin{equation}
    \sigma_k^2 = \operatorname{Var}\big[ \hat{Z}(s_0) - Z(s_0) \big]
\end{equation}
using the spatial covariance or variogram model.

\subsubsection{Original algorithm of GeoCP}

GeoConformal Prediction (GeoCP) is a model-agnostic framework for quantifying spatial prediction uncertainty by extending \emph{conformal prediction (CP)} with explicit geographic weighting.  
Conformal prediction provides finite-sample, distribution-free prediction intervals by computing nonconformity scores on a calibration set and selecting the $(1-\varepsilon)$ quantile to guarantee coverage.  
However, standard CP assumes data exchangeability and yields intervals of constant width, which is violated in geospatial settings where strong spatial heterogeneity and covariate shift are common.

To overcome these limitations, GeoCP integrates spatial dependence directly into the conformal framework.  
Given a geospatial model $f:\mathcal{X}\to\mathcal{Y}$ trained on a set of observations and a calibration set $\{(X_i,y_i)\}_{i=1}^{m}$, let $a(\cdot)$ be a nonconformity score (e.g., absolute residual) and $a_i=a(f(X_i),y_i)$ for calibration point $i$.  
For a test location $X_{\text{test}}$ with geographic coordinates $(u_{\text{test}},v_{\text{test}})$, GeoCP assigns each calibration point $i$ a geographic weight
\begin{equation}
w_i(u_{\text{test}},v_{\text{test}})=
\frac{K_\sigma \bigl(d((u_{\text{test}},v_{\text{test}}),(u_i,v_i))\bigr)}
     {\sum_{j=1}^{m} K_\sigma \bigl(d((u_{\text{test}},v_{\text{test}}),(u_j,v_j))\bigr)},
\label{eq:geo_weight}
\end{equation}
where $d(\cdot,\cdot)$ is the geographic distance and $K_\sigma$ is a distance-decay kernel (e.g., Gaussian).  
These weights reflect Tobler's first law of geography—that nearby observations are more similar—thus relaxing the exchangeability requirement of classical CP.

The GeoCP prediction interval for $X_{\text{test}}$ is then defined as
\begin{equation}
C_{\mathrm{geo}}(X_{\text{test}}) =
\left\{ y:\; a\bigl(f(X_{\text{test}}),y\bigr)\le
Q^{\mathrm{geo}}_{1-\varepsilon}\!\left(\{a_i\},\{w_i(u_{\text{test}},v_{\text{test}})\}\right)\right\},
\label{eq:geocp}
\end{equation}
where $Q^{\mathrm{geo}}_{1-\varepsilon}$ is the geographically weighted $(1-\varepsilon)$-quantile computed as
\begin{equation}
Q^{\mathrm{geo}}_{1-\varepsilon} =
\inf\left\{q:\sum_{i=1}^{m} w_i(u_{\text{test}},v_{\text{test}})\,\mathbf{1}\{a_i \le q\} \ge 1-\varepsilon \right\}.
\label{eq:geo_quantile}
\end{equation}

Algorithmically, GeoCP proceeds as follows:  
(1) split the dataset into training, calibration, and test sets;  
(2) fit the spatial prediction model $f$ on the training set;  
(3) compute nonconformity scores $\{a_i\}$ on the calibration set;  
(4) for each test point, calculate geographic weights $w_i$ via (\ref{eq:geo_weight});  
(5) determine the geographically weighted quantile (\ref{eq:geo_quantile}) and form the prediction interval (\ref{eq:geocp}).  

By construction, GeoCP inherits the rigorous finite-sample coverage guarantee of conformal prediction,
\[
\mathbb{P}\!\left[ y_{\text{test}} \in C_{\mathrm{geo}}(X_{\text{test}})\right] \ge 1-\varepsilon,
\]
while producing \emph{spatially varying} prediction intervals that directly reflect local heterogeneity.  
Because it does not require modifying the underlying predictive model, GeoCP can be applied seamlessly to classical geostatistical methods (e.g., Kriging) and modern GeoAI models, providing a unified and interpretable framework for uncertainty quantification and supporting fair, responsible geographic decision-making.

\subsection{Evolved Kriging model}

\subsubsection{GeoEvolve-Kriging (our model)}

Compared with the original Ordinary Kriging, GeoEvolve–Kriging preserves the core structure while introducing the following key innovations:

\begin{itemize}
  \item \textbf{Expanded and automatically selected variogram family.}  
  Instead of a single non-standard exponential model, GeoEvolve fits a flexible family
  \begin{equation}
  \gamma_\theta(h)=\theta_0 + \theta_1 \Bigl[1 - \exp\!\bigl(- (h/\theta_2)^{p}\bigr)\Bigr],
  \label{eq:geoevolve_variogram}
  \end{equation}
  where $p=1$ yields the exponential model, $p=2$ the Gaussian model, and $p\in(0,2)$ the \emph{Matérn} family (with smoothness $\nu$).  
  Candidate models $\{\text{Exponential},\text{Gaussian},\text{Linear},\text{Matérn}\}$ are compared using information criteria such as
  \begin{equation}
  \mathrm{AIC} = 2k - 2\log L,
  \qquad
  \mathrm{BIC} = k \log n - 2 \log L,
  \label{eq:geoevolve_modelselect}
  \end{equation}
  and the optimal variogram is selected by minimum AIC/BIC.  
  This multi-model, multi-start search avoids local minima and captures a wide spectrum of spatial smoothness.

  \item \textbf{Adaptive empirical variogram estimation.}  
  GeoEvolve constructs the empirical semi-variogram using adaptive binning based on Silverman's rule or quantiles:
  \begin{equation}
  \hat{\gamma}(h_k)
  =\frac{1}{2 |N(h_k)|}\!\!\sum_{(i,j)\in N(h_k)}\! [Z(x_i)-Z(x_j)]^2,
  \label{eq:geoevolve_empirical}
  \end{equation}
  where $N(h_k)$ is the set of pairs with distances in the $k$th adaptive bin.  
  Robust trimmed means and an automatic choice of $n_{\text{lags}}\in[8,20]\propto\sqrt{n}$ reduce the impact of outliers and distance heterogeneity.

  \item \textbf{Robust model fitting.}  
  Parameter estimation in (\ref{eq:geoevolve_variogram}) is performed via multi-start global optimization with either
  \begin{equation}
  \min_\theta \sum_k w_k \left| \hat{\gamma}(h_k)-\gamma_\theta(h_k)\right|
  \label{eq:geoevolve_loss}
  \end{equation}
  (robust L1 loss) or weighted least squares, depending on empirical residual patterns, where $w_k$ are bin-based weights.  
  This strategy guards against local minima and ensures sill $\theta_1$ and range $\theta_2$ remain physically meaningful.

  \item \textbf{Localized kriging with adaptive regularization.}  
  To improve scalability and stability, GeoEvolve restricts the kriging system to the $K$ nearest neighbors (e.g., $K=25$) of $x_0$ using a cKDTree and adds a condition-number–dependent diagonal adjustment:
  \begin{equation}
  \mathbf{K}_{\text{loc}}\lambda
  =\mathbf{k}_{\text{loc}}, \qquad
  \mathbf{K}_{\text{loc}}\leftarrow\mathbf{K}_{\text{loc}} + \epsilon(\kappa) \mathbf{I},
  \label{eq:geoevolve_local}
  \end{equation}
  where $\epsilon(\kappa)$ is an adaptive nugget (e.g., $10^{-10}$ to $10^{-4}$) determined by the matrix condition number $\kappa$.
  This reduces computational cost from $O(n^3)$ to $O(K^3)$ and stabilizes inversion in ill-conditioned settings.

  \item \textbf{Adaptive data transformation.}  
  GeoEvolve applies an adaptive log transform
  \begin{equation}
  Z^\prime = \log \bigl( Z + \delta \bigr),
  \label{eq:geoevolve_log}
  \end{equation}
  where the offset $\delta$ is chosen from the 1st percentile of positive values plus a small $\epsilon$ to reduce skewness and ensure valid back-transformation.
\end{itemize}

\subsubsection{Comparison of Evolved Kriging from Different Models}

In this section, we analyze the main technical components of different algorithm:

\textbf{Variogram family.}
Original uses only the exponential variogram with a non-standard form $nugget + sill(1 - e^{-h \cdot range})$. 
OpenEvolve standardizes the form to $e^{-h / range}$ and adds Gaussian and Linear options. 
OpenEvolve with GeoKnowledge adopts the same set but applies automatic model selection among candidate models. 
GeoEvolve further introduces the Matern family ($\nu = 0.2$--$3.0$) with full AIC/BIC-based automatic selection and multi-start optimization.

\textbf{Empirical variogram.}
Original employs 12 equal-width bins including zero distance and is unweighted. 
OpenEvolve truncates distances to 85\% of the maximum and removes NaN bins. 
OpenEvolve with GeoKnowledge follows the same procedure but adds minimal pair control. 
GeoEvolve uses adaptive binning via Silverman’s rule or quantiles, applies a robust trimmed mean, and automatically sets $n_{\text{lags}} = 8$--$20 \propto \sqrt{n}$.

\textbf{Model fitting.}
Original applies an L1 loss with a single L-BFGS-B run. 
OpenEvolve still uses L1 but adds parameter bounds, smart initialization, and a fallback strategy. 
OpenEvolve with GeoKnowledge switches to L2 loss and selects the best model by minimum MSE. 
GeoEvolve adopts a robust L1 loss, multi-start global search, Matern smoothness grid, and AIC/BIC complexity penalties.

\textbf{Kriging solver.}
Original builds a global system without neighborhood selection. 
OpenEvolve introduces diagonal regularization ($10^{-10}$) and a pseudo-inverse fallback. 
OpenEvolve with GeoKnowledge is identical. 
GeoEvolve employs localized kriging using cKDTree nearest 25 neighbors and condition-number–adaptive regularization ($10^{-10}$--$10^{-4}$), with mean fallback if the system is singular.

\subsubsection{Knowledge discovery from GeoEvolve}

We summarize the key geospatial knowledge underlying the improved GeoEvolve algorithm, which can contribute to geospatial modeling.

\textbf{Expanded variogram family with automatic selection.}  
Fits appropriate smoothness and range, lowering RMSE/MAE and improving $R^{2}$.

\textbf{Adaptive empirical variogram (trimmed mean, quantile bins).}  
Stabilizes nugget/sill/range estimates and reduces run-to-run variance.

\textbf{Multi-start with parameter bounds in optimization.}  
Improves convergence and avoids negative or degenerate parameter estimates.

\textbf{Localized kriging with condition-based regularization.}  
Reduces computational cost (from $O(n^{3})$ to local operations) and improves robustness for ill-conditioned systems.

\textbf{Geo-knowledge injection.}  
Provides informative priors and narrows the search space, improving small-sample and non-stationary performance.


\subsection{Evolved GeoCP model}
\subsubsection{GeoEvolve-GeoCP (our model)}

The fundamental conformal construction is preserved, but the following modifications are introduced:

\begin{itemize}
  \item \textbf{Refined geographic weighting.}  
  While keeping the Gaussian kernel form
  \begin{equation}
  w_i(u_{\text{test}},v_{\text{test}}) =
  \frac{\exp\!\big[-\tfrac{1}{2}\bigl(\tfrac{d((u_{\text{test}},v_{\text{test}}),(u_i,v_i))}{\sigma}\bigr)^{2}\big]}
       {\sum_{j=1}^{m} \exp\!\big[-\tfrac{1}{2}\bigl(\tfrac{d((u_{\text{test}},v_{\text{test}}),(u_j,v_j))}{\sigma}\bigr)^{2}\big]},
  \label{eq:geoevolve_weight}
  \end{equation}
  GeoEvolve reoptimizes the bandwidth parameter $\sigma$ through multi-start global search and adaptive clipping  
  \begin{equation}
  \sigma \in [\sigma_{\min},\sigma_{\max}],
  \label{eq:geoevolve_bandwidth}
  \end{equation}
  ensuring both numerical stability and fidelity to local spatial heterogeneity.

  \item \textbf{Enhanced weighted quantile computation.}  
  GeoEvolve consolidates earlier adaptive strategies into a simplified yet robust stepwise quantile estimator:
  \begin{equation}
  Q^{\mathrm{geo}}_{1-\varepsilon} =
  \inf\left\{ q : \sum_{i=1}^{m} w_i(u_{\text{test}},v_{\text{test}})\,
  \mathbf{1}\bigl\{a_i \le q \bigr\} \ge 1-\varepsilon \right\}.
  \label{eq:geoevolve_quantile}
  \end{equation}
  The algorithmic implementation uses improved vectorization and conditioning checks, guaranteeing accuracy and scalability on large test sets.
\end{itemize}

\subsubsection{Comparison of Evolved GeoCP from Different Models}

We summarize the key technical elements of the different code-evolution algorithms.

\textbf{Original GeoCP.}  
This version uses a fixed-bandwidth Gaussian kernel $e^{-0.5 d^2}$ without weight normalization. 
It computes weighted quantiles with a \emph{stepwise} rule, selecting the index where cumulative weights exceed $q$ without interpolation, and adopts the quantile level $q=\lceil (1-\alpha)(N+1)\rceil/N$, which is slightly conservative. 
Only the mean interval score is reported as the uncertainty metric. 
As a result, the method may produce overly wide or miscalibrated intervals in regions with strong spatial heterogeneity or sparse sampling.

\textbf{OpenEvolve.}  
This stage introduces adaptive bandwidth, dynamically adjusting kernel width for each test location based on its $k$-nearest neighbor distance and row-wise distance dispersion. 
It replaces the stepwise weighted quantile with interpolated weighted quantiles, avoiding discontinuous interval endpoints.

\textbf{OpenEvolve with GeoKnowledge.}  
Here the bandwidth is eo-knowledge guided: per-test $k$-NN bandwidths are clipped to the empirical range $[0.05,0.5]$. 
Weight normalization ensures that each test point’s kernel weights sum to one, providing numerical stability and spatial consistency. 
The quantile level is refined to $q=(1-\alpha)(N+1)/N$ (without ceiling), reducing conservativeness and shortening intervals. 
Furthermore, comprehensive UQ metrics are reported, including mean interval length, empirical coverage, and deviation from nominal coverage. 
Overall, this stage further shortens intervals and achieves near-nominal coverage while remaining robust at boundaries and in sparse areas.

\textbf{GeoEvolve.}  
GeoEvolve–GeoCP remains faithful to the core conformal prediction framework while sharpening spatial weighting and quantile estimation, the two pillars of interval construction.
The refined geographic weighting adaptively tunes bandwidth to local heterogeneity, ensuring that conformal scores reflect the true spatial dependence and avoid instability.

\subsubsection{Knowledge discovery from GeoEvolve}

We distill the geospatial knowledge that underlies the improved GeoCP algorithm produced by GeoEvolve.

\textbf{Adaptive bandwidth.}  
This mechanism adjusts kernel width to local calibration-point density, preventing overly wide intervals in dense regions and overly narrow ones in sparse regions. 
It drives the interval score down and keeps empirical coverage near $(1-\alpha)$.

\textbf{Interpolated weighted quantile.}  
By eliminating discrete jumps when cumulative weights cross the quantile threshold, this refinement produces smoother, more stable prediction interval endpoints and lowers variance.

\textbf{Refined quantile level without ceiling.}  
This adjustment avoids the conservative upward bias from the ceiling function, shortens interval length, and keeps empirical coverage close to the nominal level.

\end{document}